\documentclass{sig-alternate-05-2015}

\usepackage{cite}
\usepackage[caption=false,font=footnotesize]{subfig}
\usepackage{url}

\usepackage[utf8]{inputenc} % this is needed for umlauts
\usepackage[USenglish,UKenglish,english]{babel} % this is needed for umlauts
\usepackage[T1]{fontenc}    % this is needed for correct output of umlauts in pdf
\makeatletter
\newcommand{\upquotetype}{}
\newcommand{\upquote@aux}[1]{\text{\upquotetype}#1\text{\upquotetype}}
\newcommand{\upquotesingle}{\renewcommand{\upquotetype}{\textquotesingle}\upquote@aux}
\newcommand{\upquotedouble}{\renewcommand{\upquotetype}{\textquotedbl}\upquote@aux}

\usepackage{textcomp}

\usepackage{graphicx} % more modern
\usepackage{epstopdf}
\usepackage{amsfonts}
\usepackage{amsmath}

\usepackage{algorithm}
\usepackage{algorithmic}

\makeatletter
\global\copyrightetc{
{\noindent\confname\ \the\conf\ \the\confinfo}\par\smallskip
  \if@printcopyright
    \copyright\ \the\copyrtyr\ \@copyrightowner
  \fi
} 
\toappear{\fontsize{7pt}{8pt}\fontfamily{ptm}\selectfont
  \the\boilerplate\par\smallskip
 \the\copyrightetc}
\makeatother

\begin{document}

% Copyright
%\setcopyright{acmcopyright}
%\setcopyright{acmlicensed}
%\setcopyright{usgov}
%\setcopyright{usgovmixed}
%\setcopyright{cagov}
%\setcopyright{cagovmixed}

\setcopyright{rightsretained}

% DOI
%\doi{10.475/123_4}

% ISBN
%\isbn{123-4567-24-567/08/06}

%Conference
\conferenceinfo{KDD 2016 Workshop on Interactive Data Exploration and Analytics (IDEA'16)}{August 14th, 2016, San Francisco, CA, USA.}

%\acmPrice{\$15.00}

\CopyrightYear{2016}
%\copyrightetc{Copyright is held by the owner/author(s)}

%
% --- Author Metadata here ---
%\CopyrightYear{2007} % Allows default copyright year (20XX) to be over-ridden - IF NEED BE.
%\crdata{0-12345-67-8/90/01}  % Allows default copyright data (0-89791-88-6/97/05) to be over-ridden - IF NEED BE.
% --- End of Author Metadata ---

\title{Interacting with Massive Behavioral Data}

\numberofauthors{1} %  in this sample file, there are a *total*
\author{
% 1st. author
\alignauthor
Shih-Chieh Su\\
       \affaddr{Qualcomm Inc.}\\
       \affaddr{5775 Morehouse Drive}\\
       \affaddr{San Diego, CA 92121 USA}\\
       \email{shihchie@qualcomm.com}
}
% There's nothing stopping you putting the seventh, eighth, etc.
% author on the opening page (as the 'third row') but we ask,
% for aesthetic reasons that you place these 'additional authors'
% in the \additional authors block, viz.
\date{30 July 1999}
% Just remember to make sure that the TOTAL number of authors
% is the number that will appear on the first page PLUS the
% number that will appear in the \additionalauthors section.

\maketitle
\begin{abstract}
In this short paper, we propose the split-diffuse (SD) algorithm that takes the output of an existing word embedding algorithm, and distributes the data points uniformly across the visualization space. The result improves the perceivability and the interactability by the human.

We apply the SD algorithm to analyze the user behavior through access logs within the cyber security domain. The result, named the topic grids, is a set of grids on various topics generated from the logs. On the same set of grids, different behavioral metrics can be shown on different targets over different periods of time, to provide visualization and interaction to the human experts.

Analysis, investigation, and other types of interaction can be performed on the topic grids more efficiently than on the output of existing dimension reduction methods. In addition to the cyber security domain, the topic grids can be further applied to other domains like e-commerce, credit card transaction, customer service to analyze the behavior in a large scale.
\end{abstract}

\begin{CCSXML}
<ccs2012>
<concept>
<concept_id>10003120.10003145.10003147.10010365</concept_id>
<concept_desc>Human-centered computing~Visual analytics</concept_desc>
<concept_significance>500</concept_significance>
</concept>
<concept>
<concept_id>10003120.10003145.10003147.10010923</concept_id>
<concept_desc>Human-centered computing~Information visualization</concept_desc>
<concept_significance>500</concept_significance>
</concept>
</ccs2012>
\end{CCSXML}

\ccsdesc[500]{Human-centered computing~Visual analytics}
\ccsdesc[500]{Human-centered computing~Information visualization}

%
%  Use this command to print the description
%
\printccsdesc

% We no longer use \terms command

\keywords{data visualization, human interaction, dimension reduction, risk management}

\section{Introduction}

When there are multiple measures of the each sample, the data is described in the a high dimensional space $\mathcal{H}$ by these measures. To make these high dimensional data points visible to human, a word embedding (or dimension reduction) technique is employed to map the data points to a lower dimensional space $\mathcal{L}$. Usually $\mathcal{L}$ is a two-dimensional (2D) or three-dimensional (3D) space. The word embedding technique of choice attempts to preserve some relationship among the data points in $\mathcal{H}$ after mapping them to $\mathcal{L}$. 

For example, the multi-dimensional scaling (MDS) \cite{Torgerson52} $\mathbb{M}$ tries to preserve the distance between data points, during the mapping from $\mathcal{H}$ to $\mathcal{L}$. The stochastic neighbor embedding (SNE) \cite{Van08} type of algorithms further emphasize the local relationship ahead of the global relationship. There are other dimension reduction techniques putting emphasis on different favored metrics over relationship. On a specific situation, one particular dimension reduction technique could be more suitable or more efficient than others. 

The output from existing dimension reduction algorithm is a set of data points that are non-uniformly scattered around the visualization space, which has some drawbacks:
\begin{enumerate}
  \item Some data points may overlap with others. Overlap makes the information less perceivable.
  \item The data points are denser in some area. The heterogeneity makes human interaction with the data points more difficult.
\end{enumerate}

\begin{figure*}[htp]
  \centering
  \subfloat[2D t-SNE output]{\includegraphics[scale=0.14]{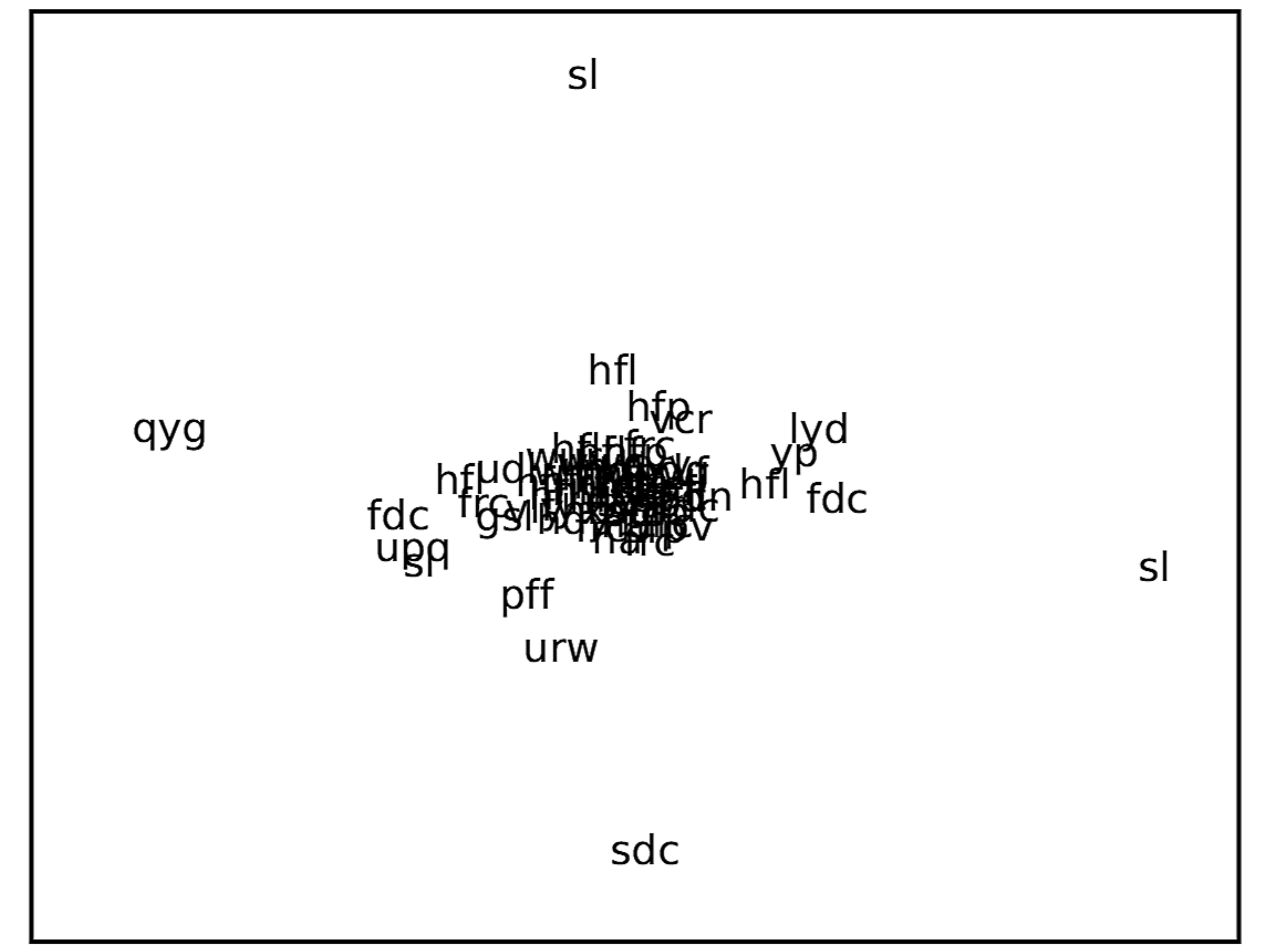}}
  \subfloat[2D MDS output]{\includegraphics[scale=0.14]{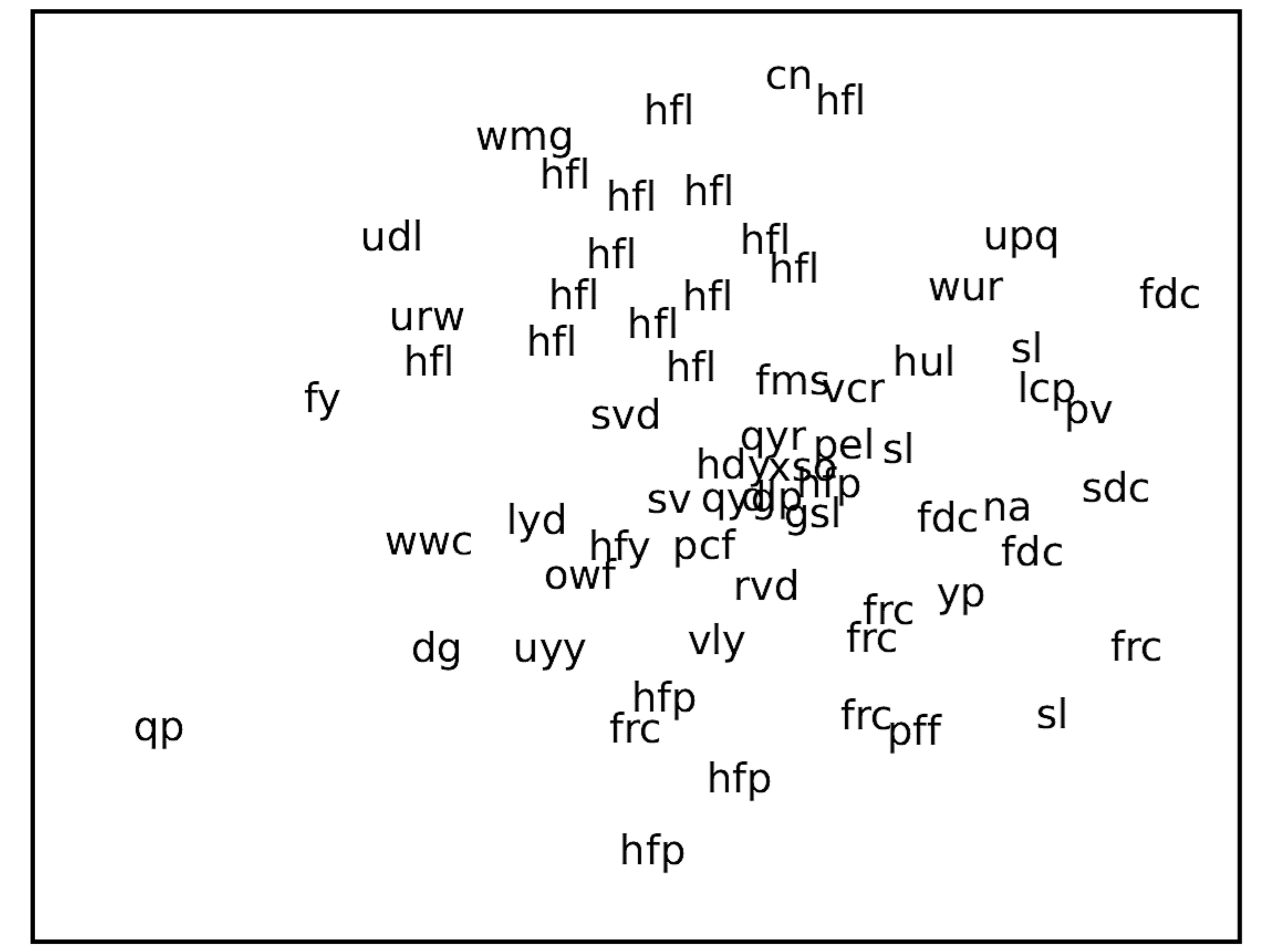}}
  \quad
  \subfloat[3D t-SNE output]{\includegraphics[scale=0.14]{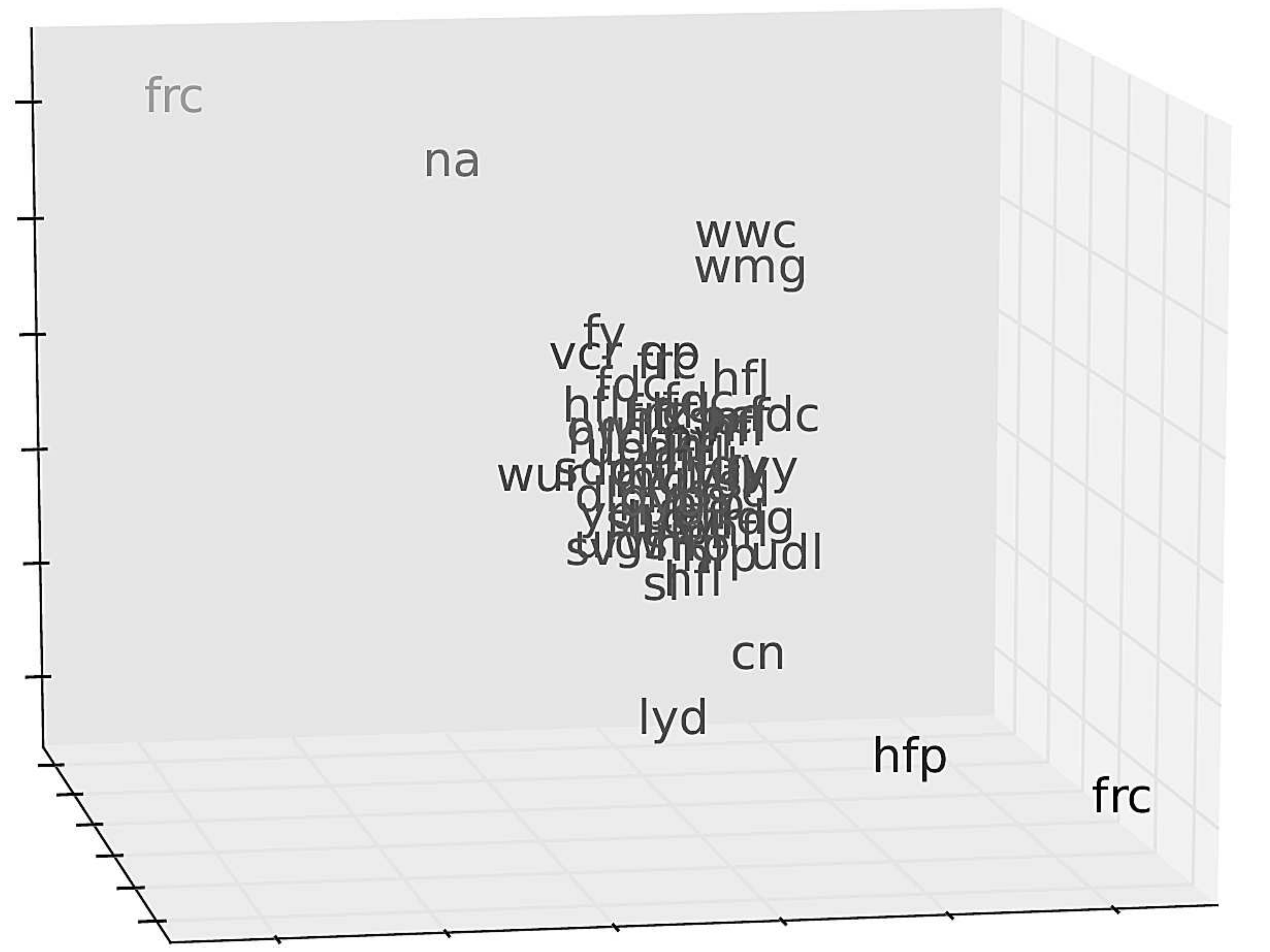}}
  \subfloat[3D MDS output]{\includegraphics[scale=0.14]{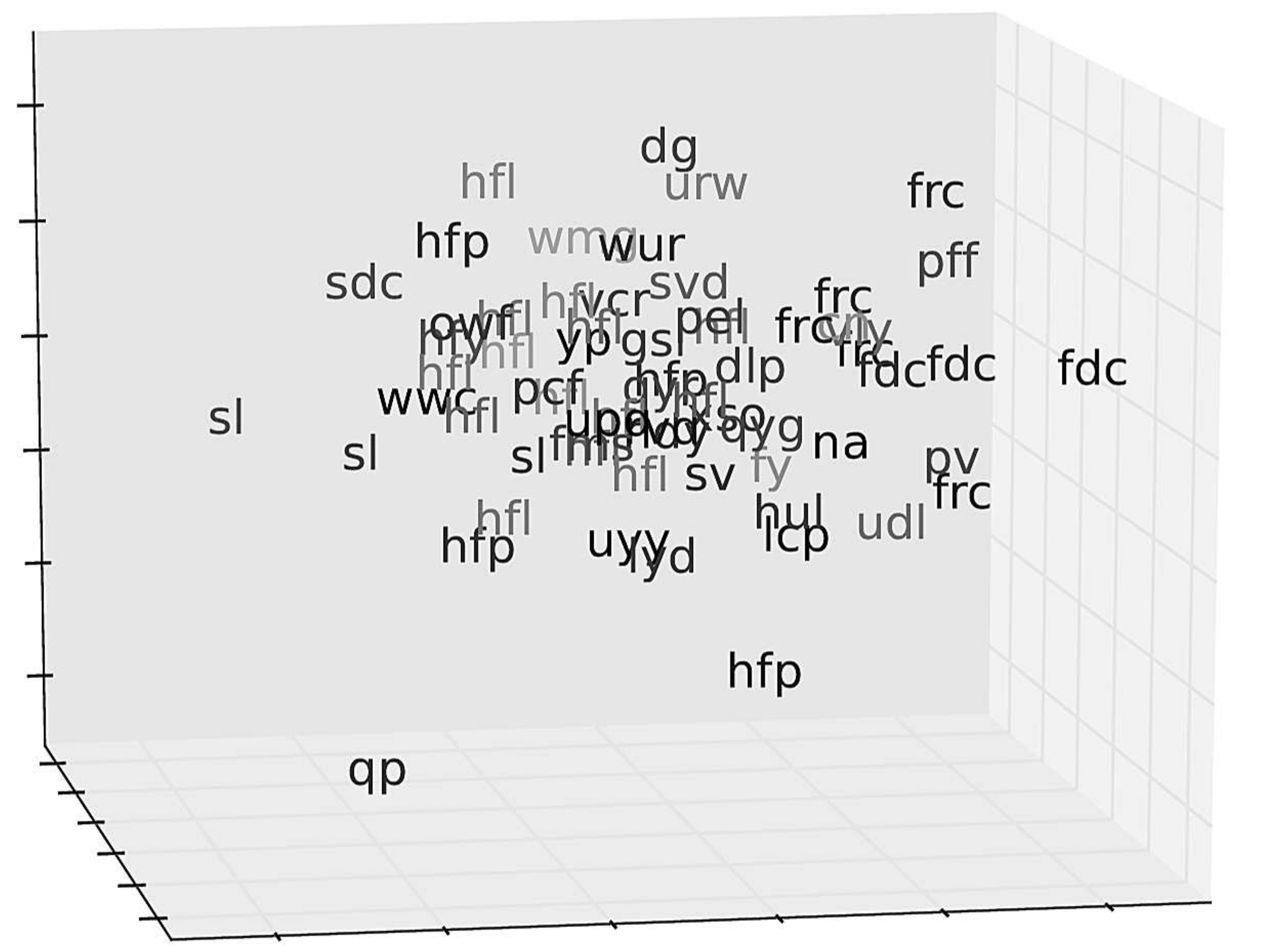}}
  \\
  \subfloat[SD output from (a)]{\includegraphics[scale=0.14]{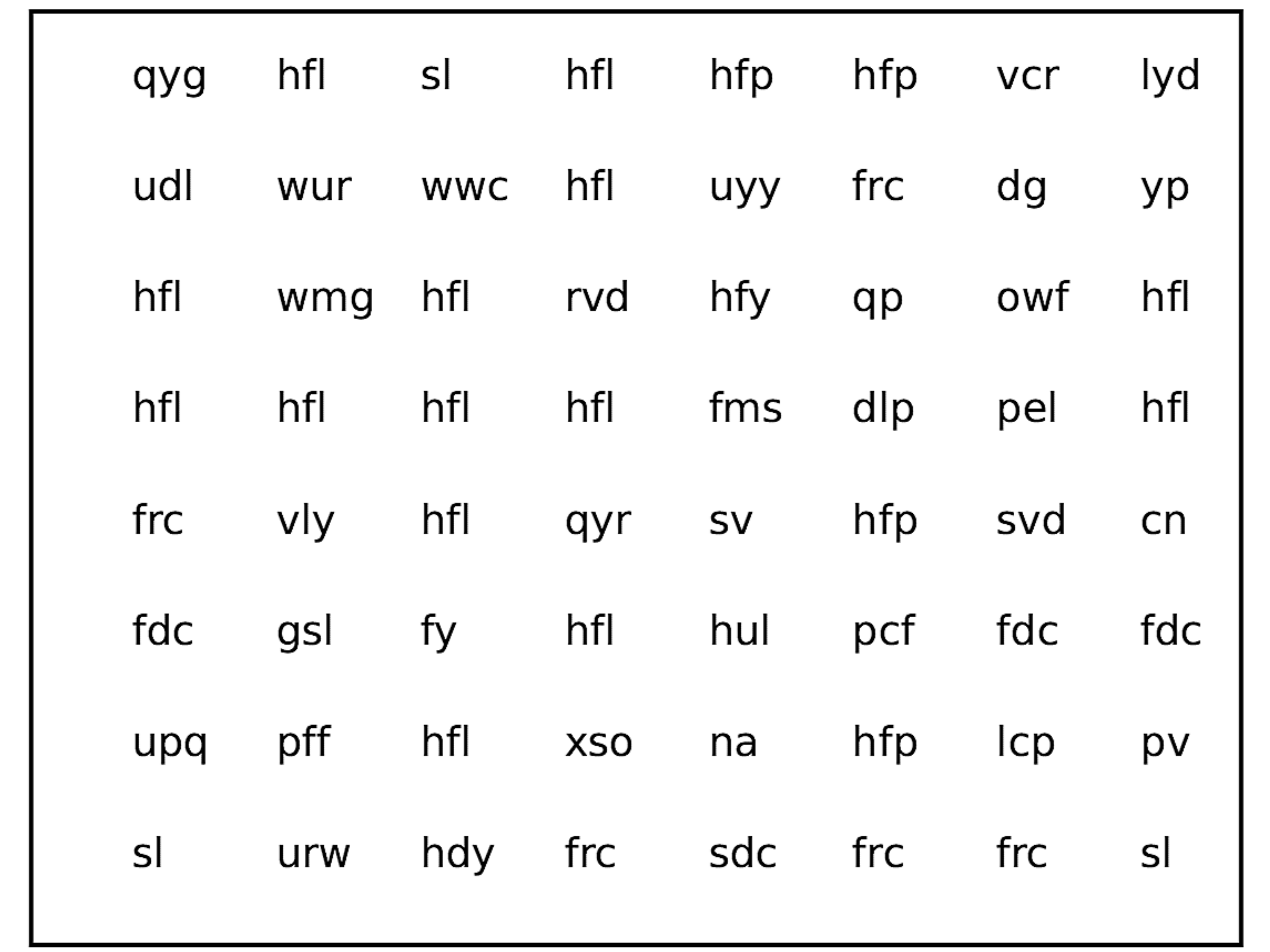}}
  \subfloat[SD output from (b)]{\includegraphics[scale=0.14]{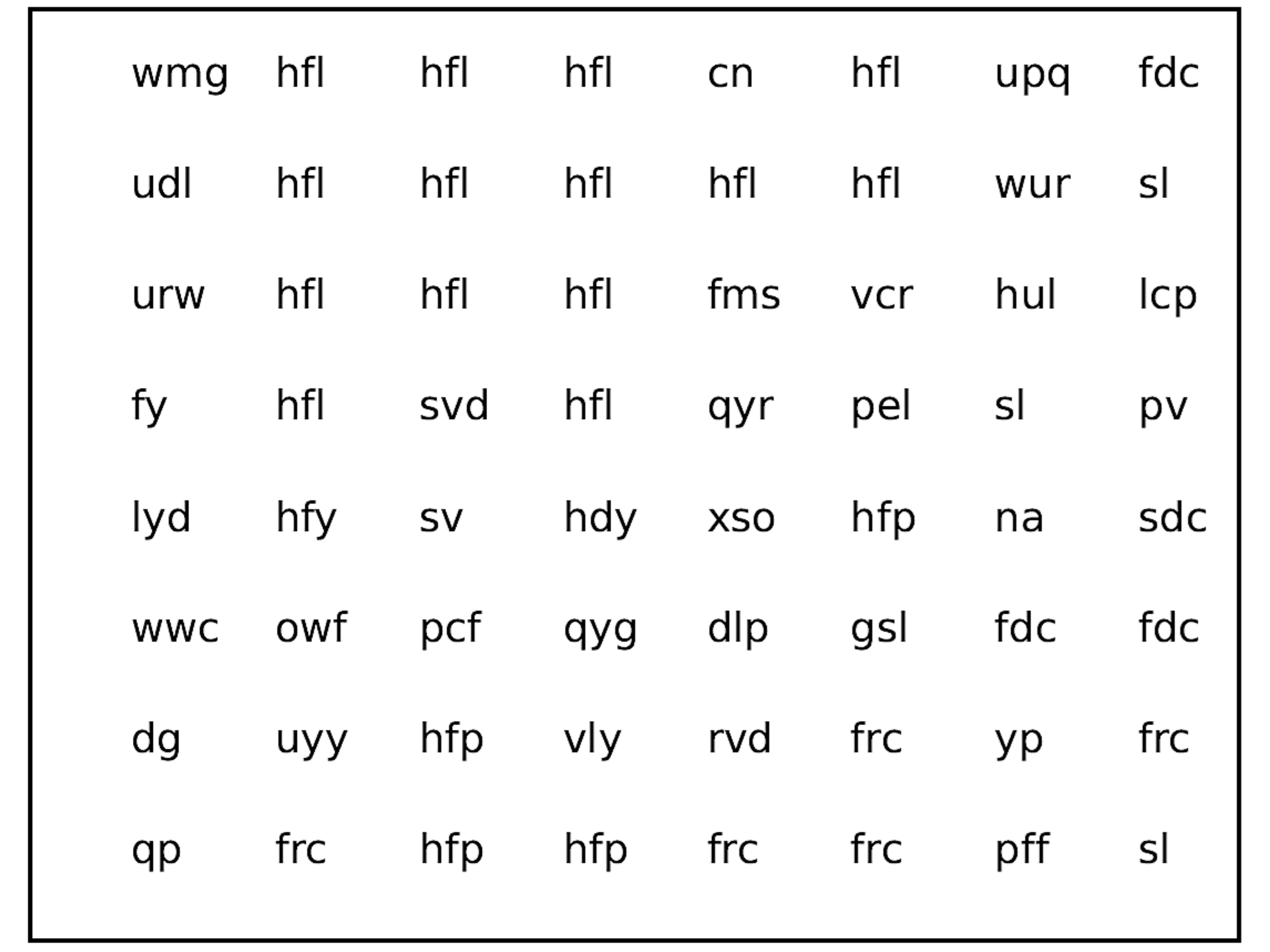}}
  \quad
  \subfloat[SD output from (c)]{\includegraphics[scale=0.14]{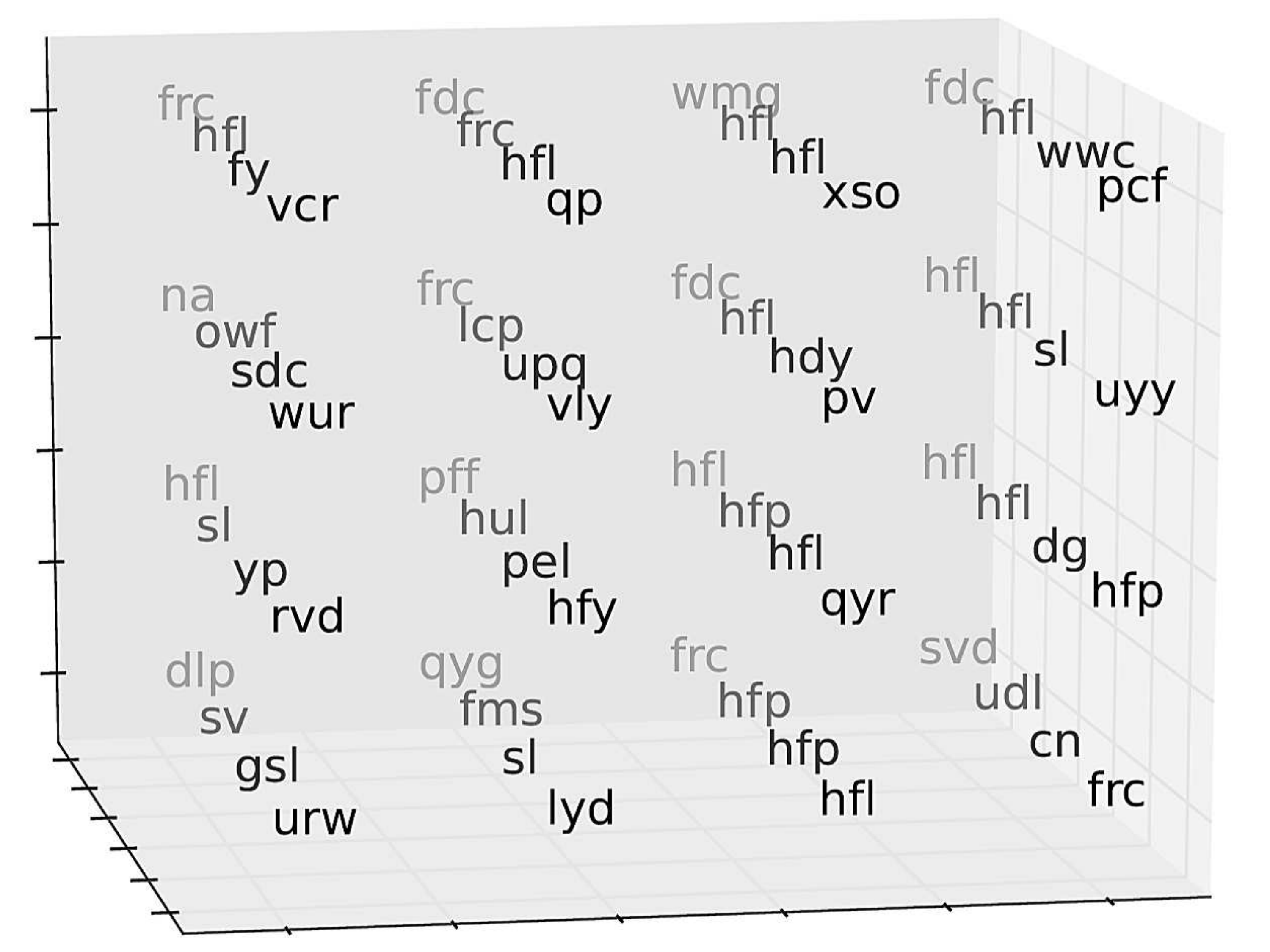}}
  \subfloat[SD output from (d)]{\includegraphics[scale=0.14]{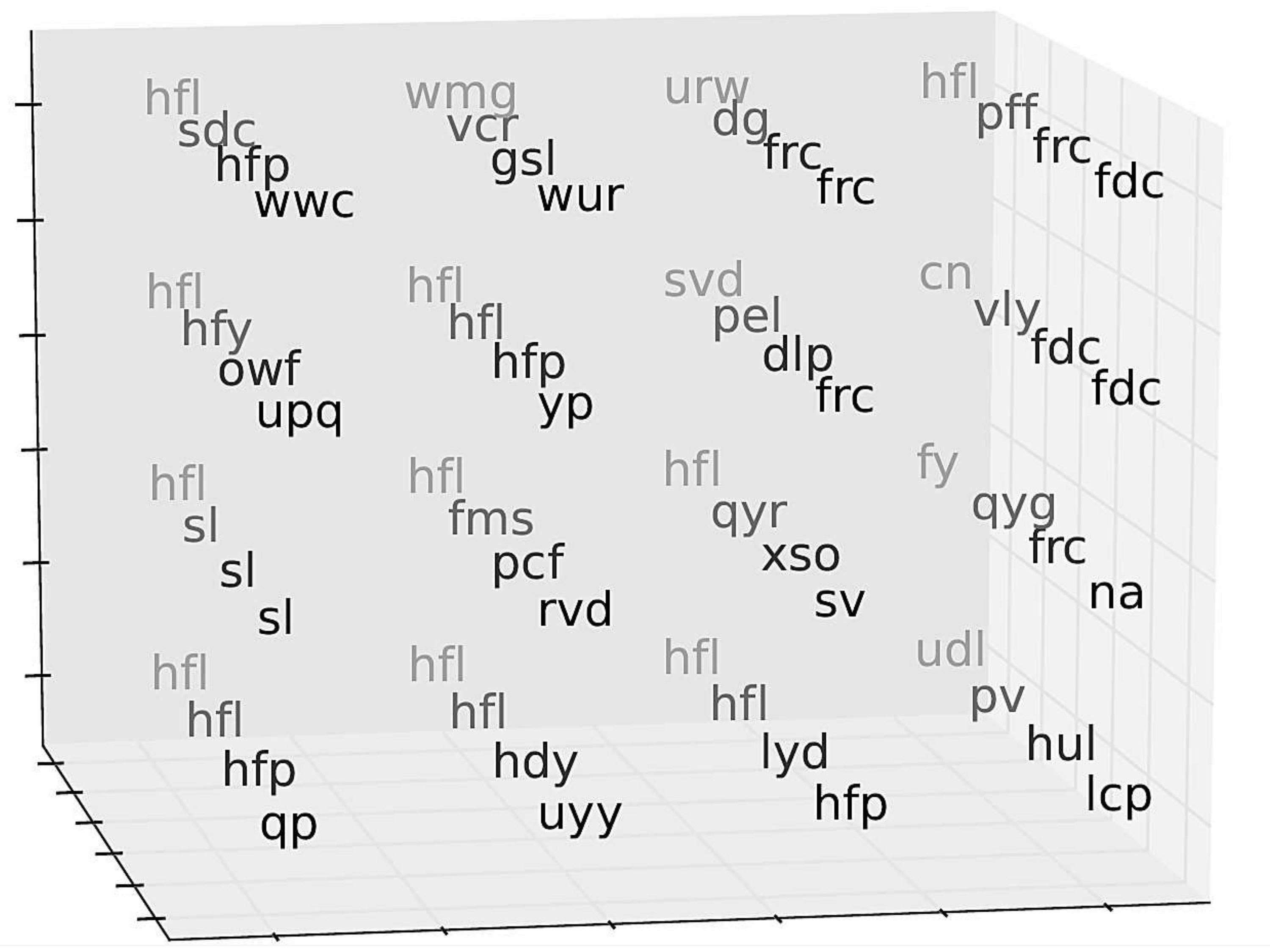}}
\caption{The split-diffuse (SD) algorithm takes the output of any dimension reduction technique and distributes the data points evenly while maintaining the topology among them. Example inputs of 64 data points from (a) 2D t-SNE, (b) 2D MDS, (c) 3D t-SNE and (d) 3D MDS are distributed evenly in the same space as shown in (e)-(h) respectively.}
\label{fig-teaser}
\end{figure*}

\section{Methods}

In order to better utilize the visualization space, we proposed to distribute the data points evenly over the visualization space. The cloud of data points is deformed in the same space defined by the dimension reduction algorithm of choice. This deformation is denoted as $\mathbb{S}$. In the meanwhile, it is desirable to preserve the point-wise relationship maintained by the dimension reduction algorithm. Our strategy in approaching this goal is prioritized as follows:
\begin{enumerate}
  \item Points are equally spaced after the mapping $\mathbb{S}$.
  \item Point-wise topology is preserved. $\mathbb{S}$ attempts to keep point $p_j$ on the same side of point $p_i$ as before the mapping.
  \item Point-wise geometry is loosely followed. When $p_i$ is far from $p_j$, $\mathbb{S}(p_i)$ is far from $\mathbb{S}(p_j)$.
\end{enumerate}

\begin{algorithm}[tb]
   \caption{Split-diffuse algorithm (square of power of $2$)}
   \label{alg-SD}
   {\bfseries Input:} data points $\{p\}$ of length $2^h \times 2^h$, depth $d=0$, allocation string $c=\upquotesingle{ }$\\ 
   {\bfseries split-diffuse ($\{p\}$, $d$, $c$)}
\begin{algorithmic}
   \STATE $k \leftarrow $ length of $\{p\}$
   \IF{$k = 1$,} 
   \STATE resolve $\mathbb{S}(p)$ from $c$
   \STATE return $p$
   \ENDIF
   \STATE $a \leftarrow mod($depth$,2)$
   \STATE $m \leftarrow $ median of $\{p\}$ in the dimension $a$
   \STATE 
   \begin{tabular}{@{}ll@{}l} 
   return & ( & [{\bfseries split-diffuse} ($\{p:p\leq m|_{dim=a}\}$, $d$+1, $c$+\upquotesingle{L})], \\ 
   & & [{\bfseries split-diffuse} ($\{p:p> m|_{dim=a}\}$, $d$+1, $c$+\upquotesingle{R})])
   \end{tabular}
\end{algorithmic}
\end{algorithm}

\begin{figure}[htp]
  \centering
  \subfloat[first-level split]{\includegraphics[scale=0.15,angle=-90]{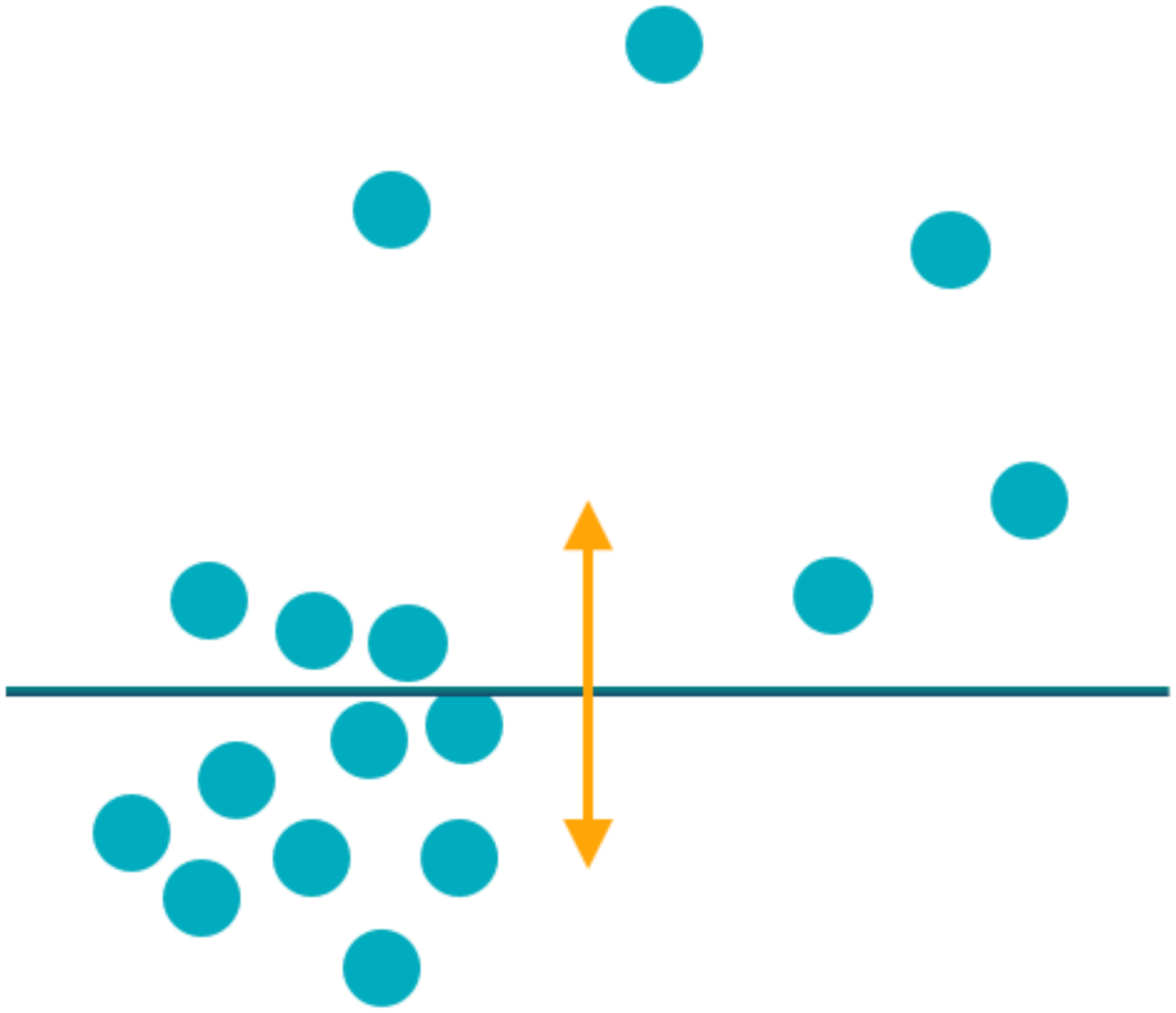}}
  \subfloat[second-level splits]{\includegraphics[scale=0.15,angle=-90]{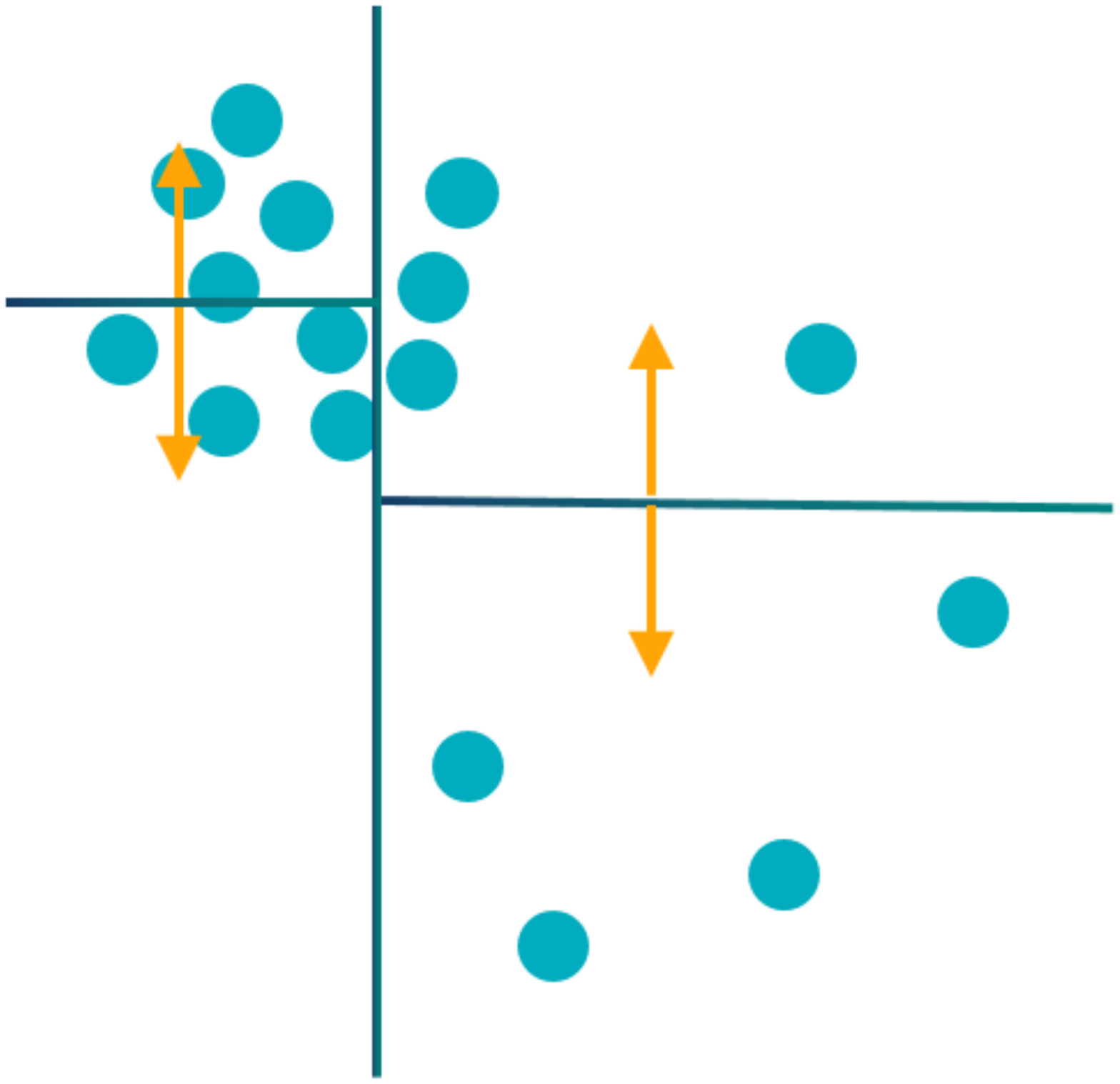}}
\caption{The split-diffuse algorithm over the $4\times4$ layout.}
\label{fig-SD}
\end{figure}

The algorithm we propose is called the split-diffuse (SD) algorithm (Algorithm~\ref{alg-SD}), which follows the strategies above. In our implementation, the SD algorithm first picks the $x$-axis as the dimension to split. As in Figure~\ref{fig-SD} (a), it splits the data points into two groups: the ones smaller than or equal to the median, and the ones larger than the median. Each group goes through this split step again over the $y$-dimension, as in Figure~\ref{fig-SD} (b). We recursively split the points in $x$- and $y$-dimension iteratively, until there is only one point in current recursion.

We keep track of the splitting path in string $c$. At the end of the recursion, the placement each single point $p$ is resolved. The indexes of the SD-mapped points, $\mathbb{S}(p)$, are all integers, and forms a $2^h \times 2^h$ array. This means that the mapped data points are equally spaced in a $2^h \times 2^h$ square. To achieve this uniformity in the space $\mathcal{L}$, the data points are essentially diffused from the denser area to the coarser area by the SD algorithm --- hence the name split-diffuse.

Some sample outputs from existing dimension reduction techniques are shown in Figure~\ref{fig-teaser}, as well as the corresponding SD outputs. Although we only present the results from t-SNE and MDS, the SD algorithm can be applied to outputs of other techniques such as the principal component analysis (PCA) \cite{Pearson01}, isomap \cite{Tenenbaum00}, spectral embedding \cite{Belkin03}, and totally random trees embedding \cite{Geurts06}.

\begin{figure*}[htp]
  \centering
  \subfloat[current activities of an entity]{\includegraphics[scale=0.16]{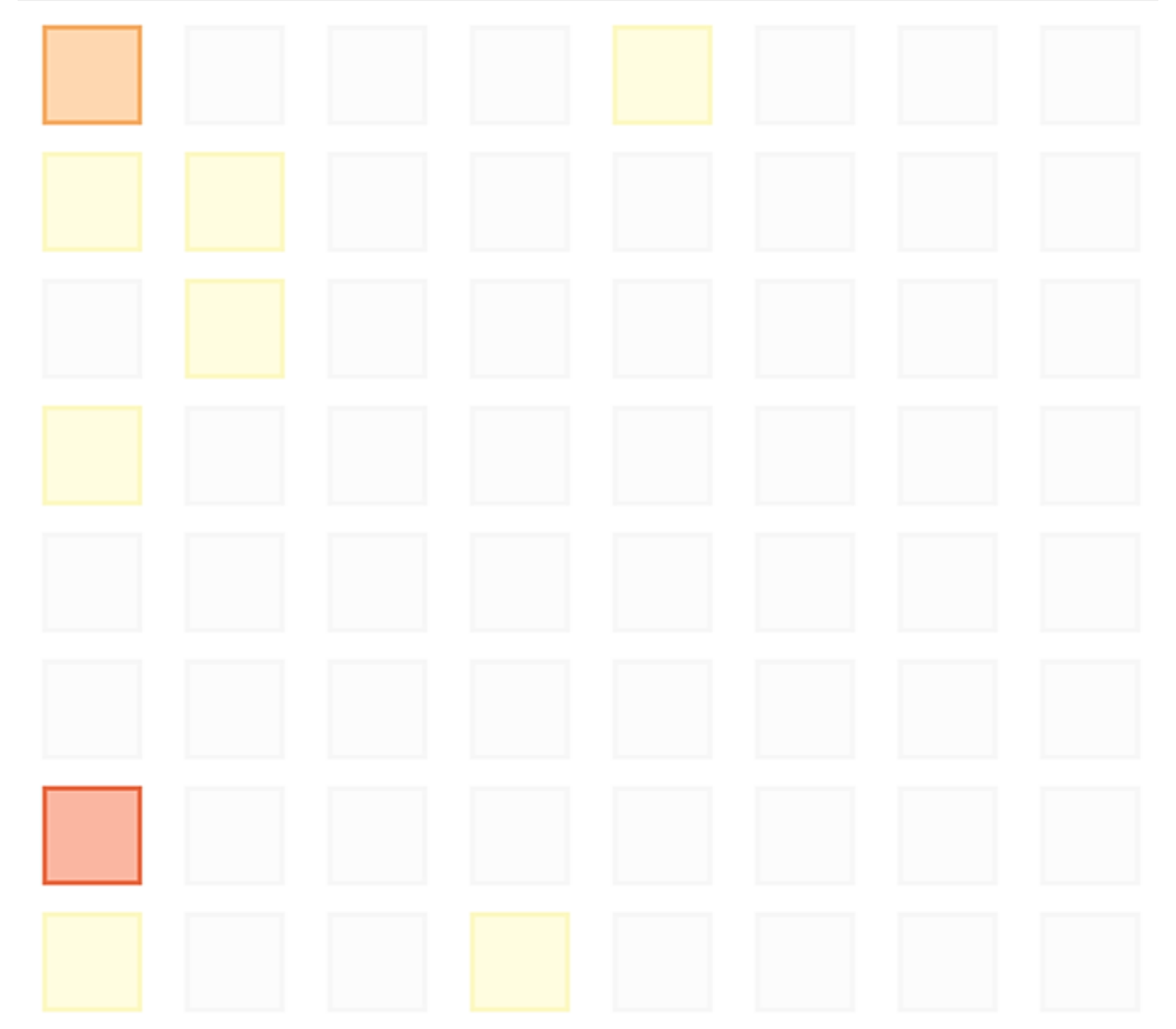}}\quad
  \subfloat[historical activities of this entity]{\includegraphics[scale=0.16]{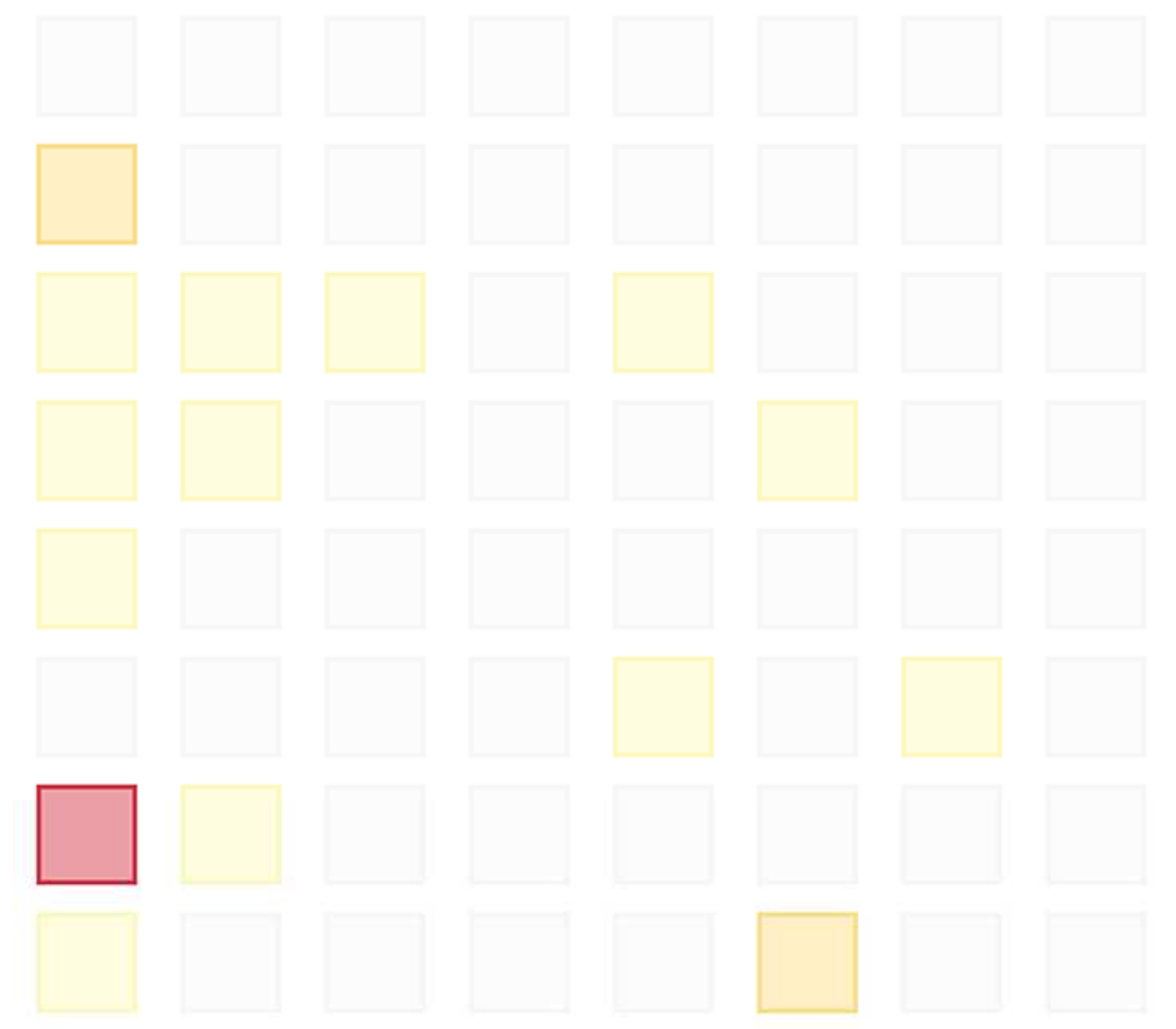}}\quad
  \subfloat[risk against the historical activities of this entity]{\includegraphics[scale=0.16]{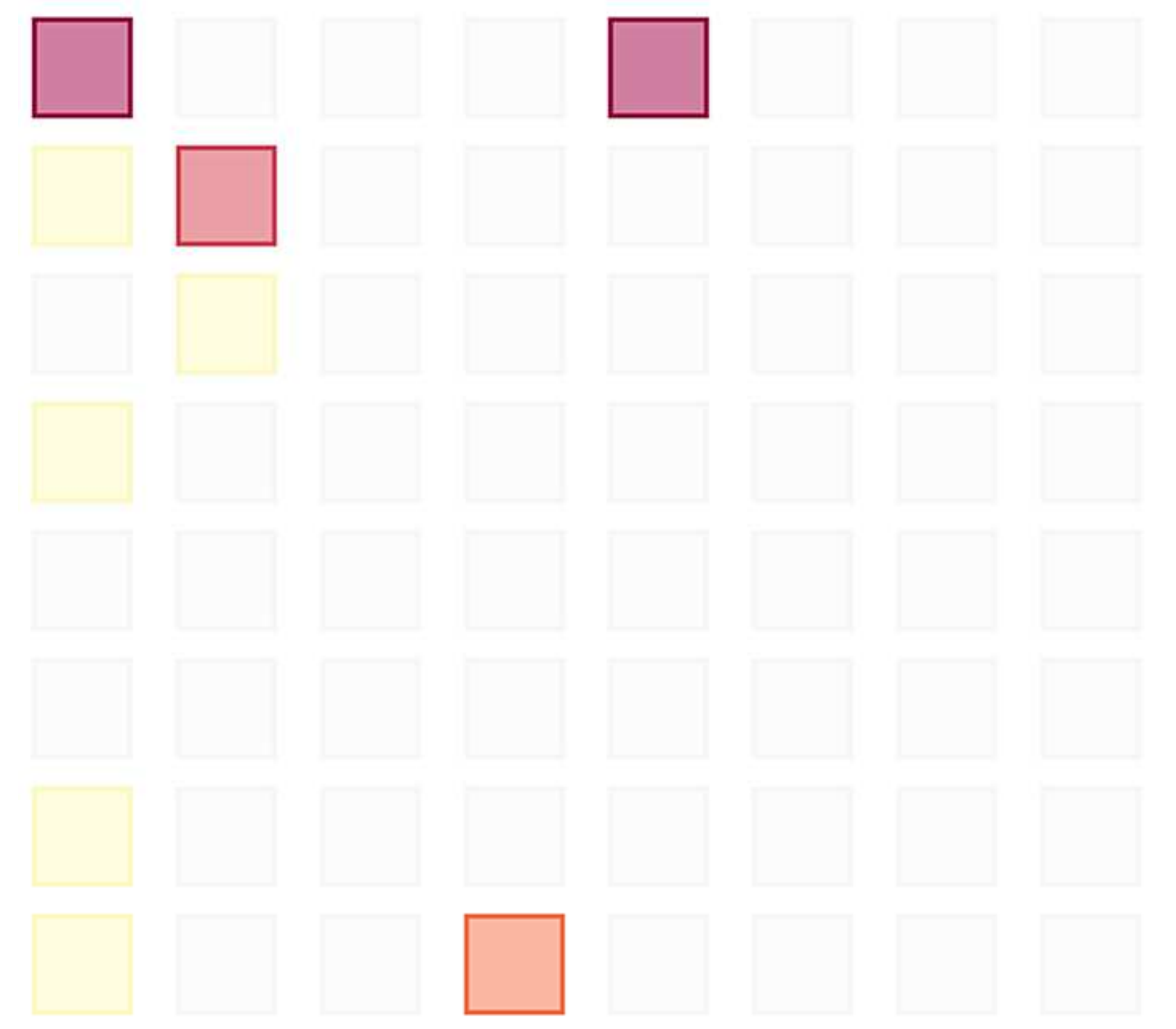}}\quad
  \subfloat[historical activities of the peers]{\includegraphics[scale=0.16]{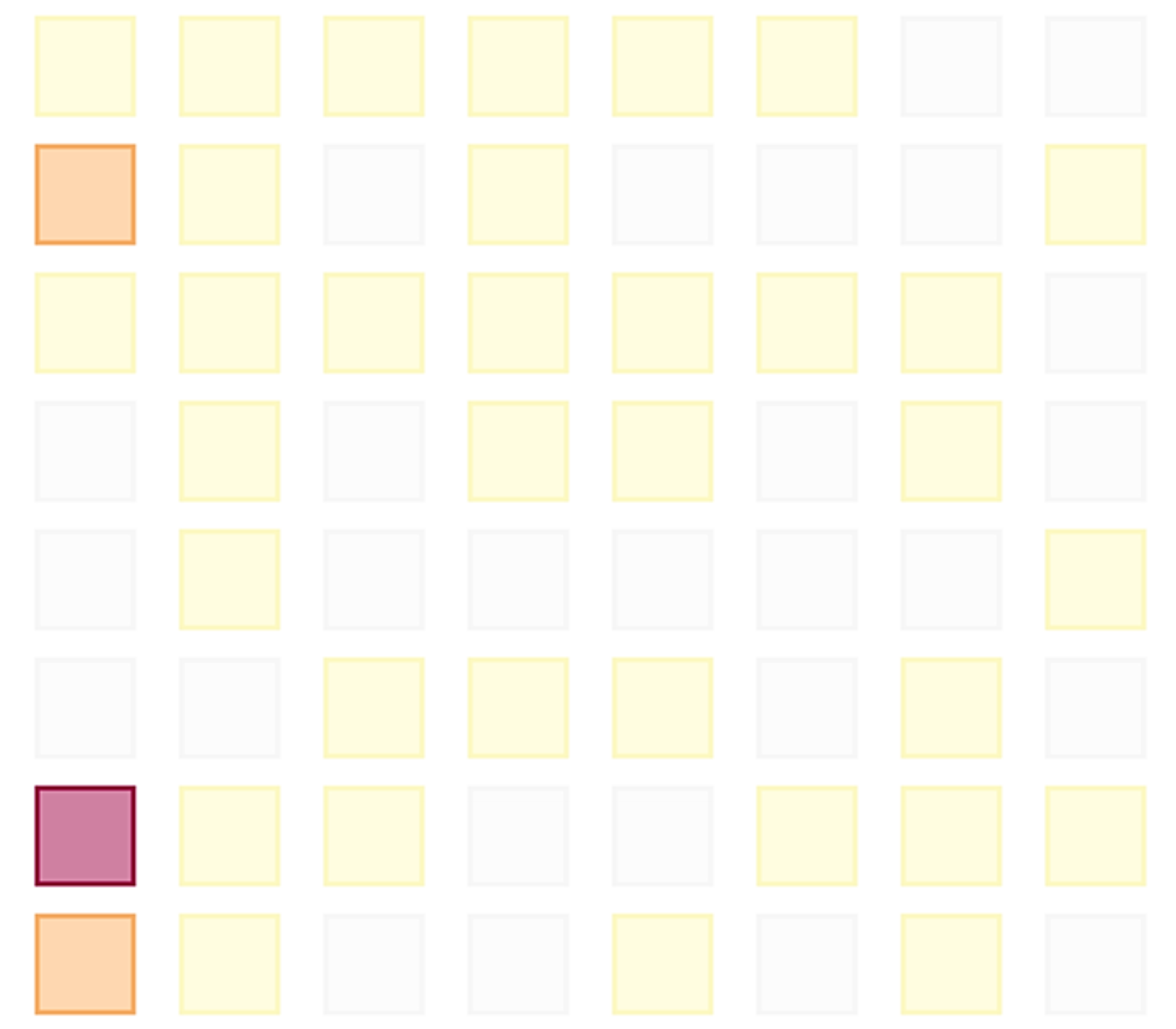}}\quad
  \subfloat[risk against the historical activities of the peers]{\includegraphics[scale=0.16]{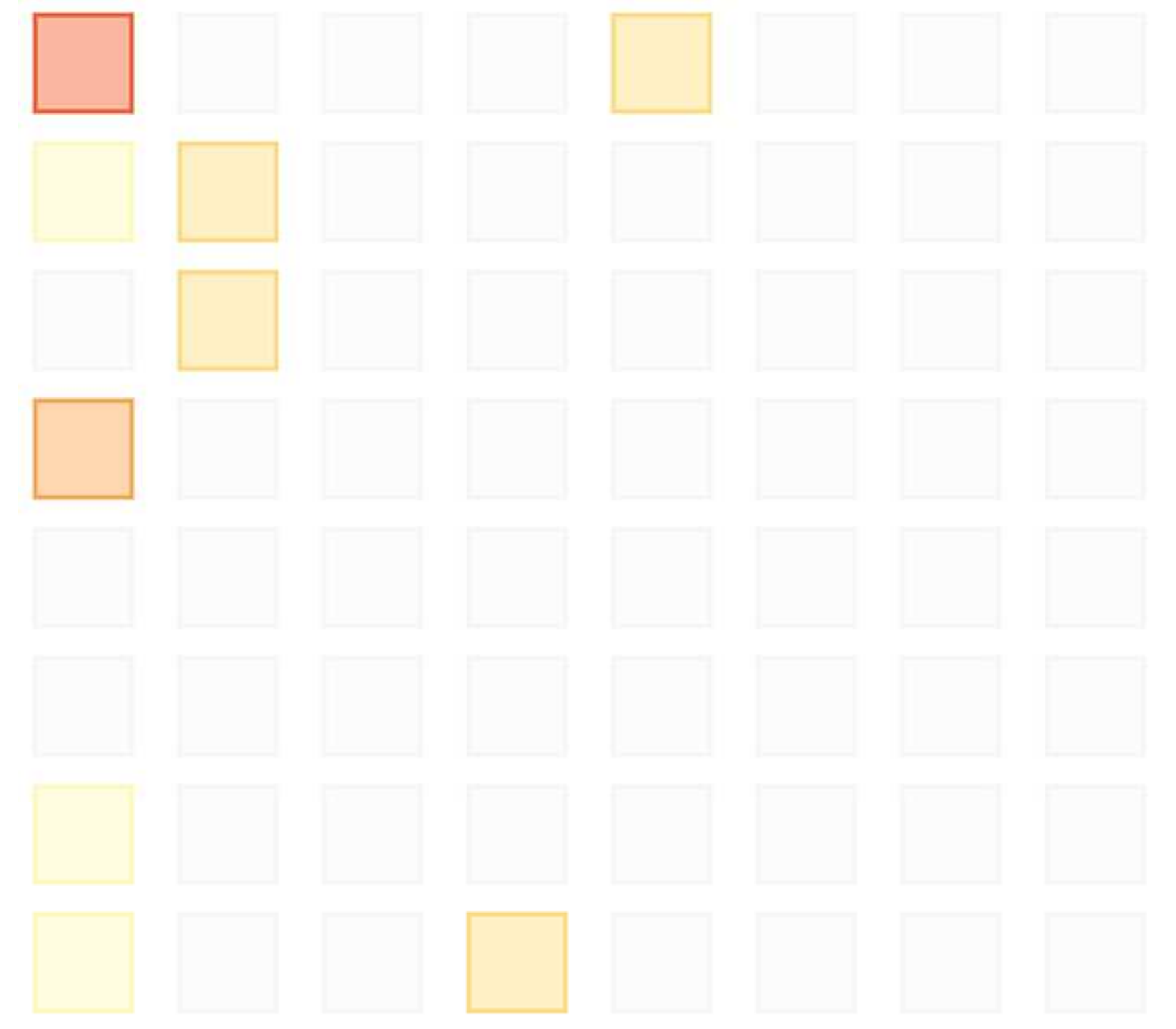}}
\caption{The topic grids. The self risk in (c) is derived from comparing the current activities (a) and the historical activities (b) of a specific entity. The peer risk in (e) is derived from comparing the current activities (a) and the peers' activities (d) of a specific entity.}
\label{fig-TG}
\end{figure*}

%\section{Application}
\section{Interacting with the Data}

The motivation to better utilize the visual space comes from the need to interact with massive amount of data. Consider the case that there are millions of items being shipped to a city every month. How can we easily observe the difference in the monthly shipping patterns? When vectoring each item as a data point, putting all the data points on a chart makes the chart hard to read. Instead, using clustering algorithms to group the points and showing the representatives is a better way to present the shipping pattern. Still, with the existing dimension reduction techniques (Figure~\ref{fig-teaser} (a)-(d)), it is difficult to visually compare the difference and interact with the representative points for more detail.

In our use case, we apply the SD algorithm to help analyzing behavioral content in the cyber security domain. The goal of the system is to detect behavioral anomaly based on the access logs. Topics are generated on the content of the logs in a word vector space of $19K+$ dimensions. MDS is applied to reduce the dimension. As shown in Figure~\ref{fig-teaser} (b) and (f), topics are represented by the most relevant keywords, encrypted. Topics close to each other may share the same representative keyword. The SD algorithm follows to generate the topic grids and visualize different metrics about the behavior of a user (Figure~\ref{fig-TG}).

When not directly displaying the detail keywords about a topic, the topic grids requires less space. At the same time, the human expert still can easily keep track of the topics based on their indexes over all dimensions and compare the difference between different sets of topic grids. Human interaction, which is the ultimate goal of the uniform placement of the data points, can be done more easily on the topic grids than on the raw dimension reduction output as in Figure~\ref{fig-teaser} (a)-(d). For example, the mouse over event on a grid pops up the topical summary, and the click event to overlay the detailed topical activities.

\begin{figure}[htp]
  \centering
  \subfloat[Topic curtain]{\includegraphics[scale=0.35]{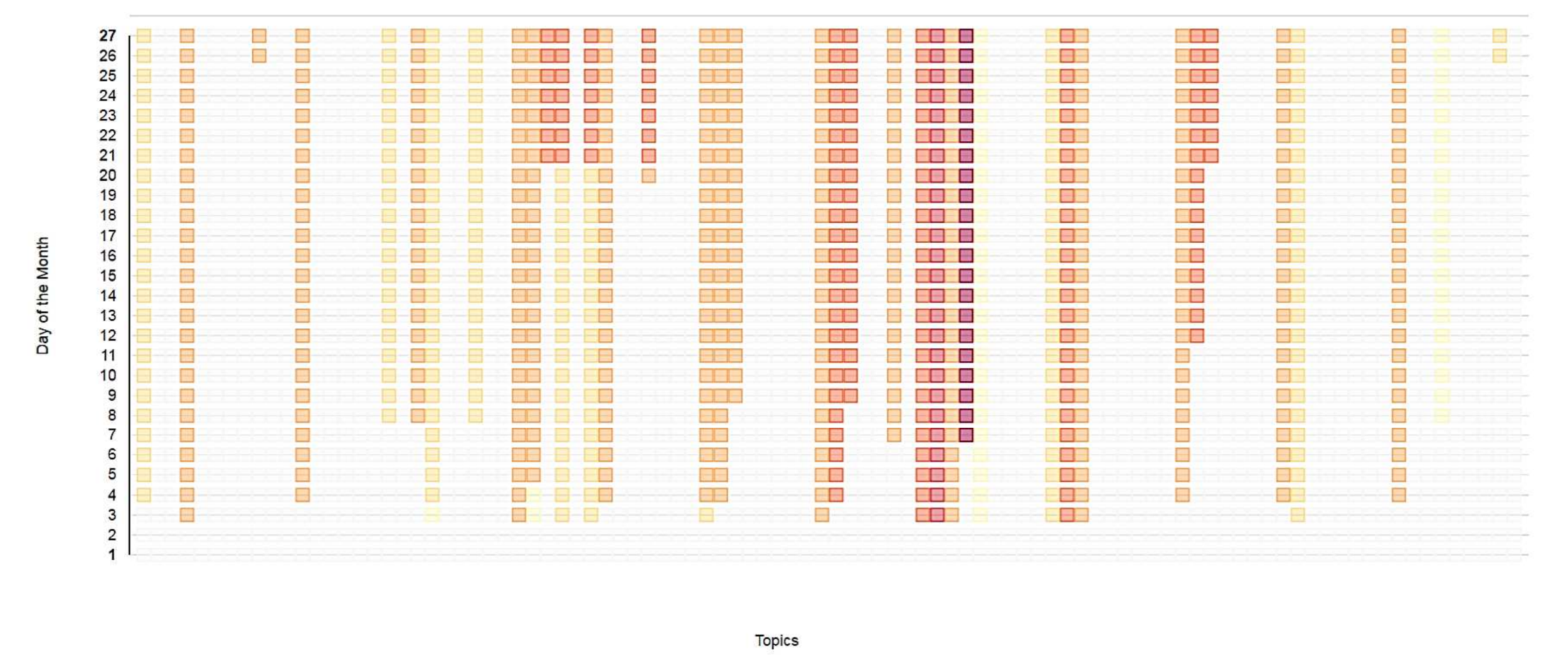}}\\
  \subfloat[Topic shower]{\includegraphics[scale=0.22]{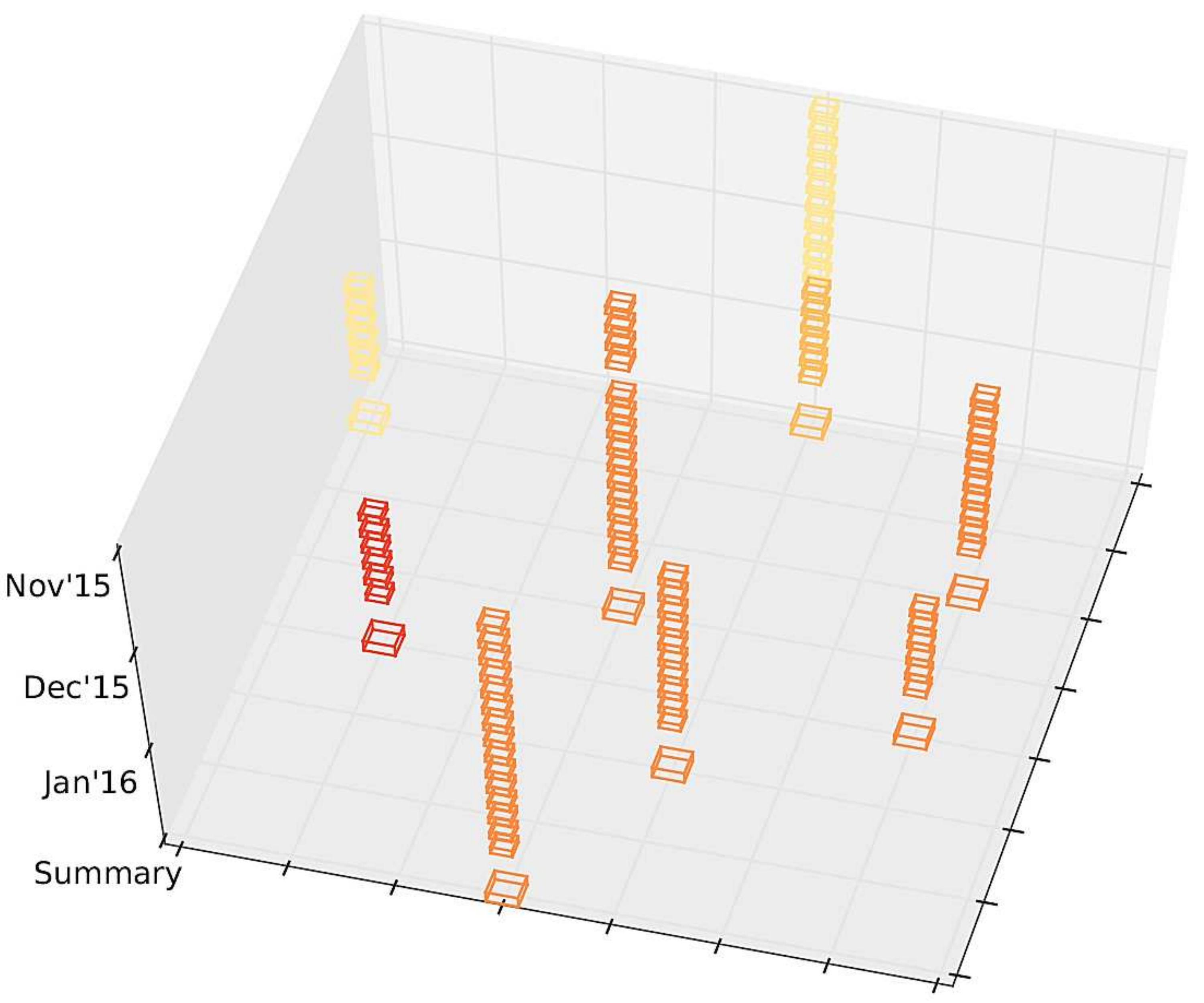}}
\caption{Other formats of the topic grids}
\label{fig-TCS}
\end{figure}

It is also useful to monitor the behavior change over time. In such cases, we reserve a dimension in $\mathcal{L}$ as the time axis. For a 2D space $\mathcal{L}$, a 1D version of SD algorithm is applied to maintain the point-wise topology. The cumulative activities have a  shape of curtain. Meanwhile, we can pile up the 2D topic grids on the time axis over the 3D $\mathcal{L}$, as shown in Figure~\ref{fig-TCS}. With normal or usual behavior, it is expected to see the consistent hot grids at the same locations over time.

\section{Future Work}

In addition to the cyber security domain, the topic grids can be applied to other domains having free-form text logs to analyze the behavior described by the logs. Some possible use cases include e-commerce, credit card transaction, customer service, or others with large volume of behavioral data to be analyzed.

It is also possible to apply the topic grids to the structured data, on which an arbitrary clustering algorithm can generate cluster centers. The data points are then organized into these cluster centers, the same way we use the topic to represent the log entries related to it.

%\bibliography{template}
\bibliographystyle{abbrv}
\bibliography{jsu2016}

\end{document}